\documentclass{article}

\usepackage{arxiv}

\usepackage[utf8]{inputenc} 
\usepackage[T1]{fontenc}    
\usepackage{hyperref}       
\usepackage{url}            
\usepackage{booktabs}       
\usepackage{amsfonts}       
\usepackage{nicefrac}       
\usepackage{microtype}      
\usepackage{lipsum}
\usepackage{graphicx}
\usepackage{amsmath}
\graphicspath{ {./images/} }

\title{An Explainable Machine Learning Approach for Age and Gender Estimation in Living Individuals Using Dental Biometrics}

\author{
Mohsin Ali \\
University of Essex, UK \\
\texttt{ma22159@essex.ac.uk} \\
\And
Haider Raza \\
University of Essex, UK \\
\texttt{h.raza@essex.ac.uk} \\
\And
John Q Gan \\
University of Essex, UK \\
\texttt{jqgan@essex.ac.uk} \\
\And
Ariel Pokhojaev \\
Tel Aviv University, Israel \\
\texttt{nikitap@mail.tau.ac.il} \\
\And
Matanel Katz \\
Tel Aviv University, Israel \\
\texttt{matanelkatz@mail.tau.ac.il} \\
\And
Esra Kosan \\
Oral Medicine and Oral Surgery, Germany \\
\texttt{esra-celin.kosan@charite.de} \\
\And
Dian Agustin Wahjuningrum \\
Universitas Airlangga, Indonesia \\
\texttt{dian-agustin-w@fkg.unair.ac.id} \\
\And
Omnina Saleh \\
Boston University Henry M.Goldman, USA \\
\texttt{omnia99.ismail@gmail.com} \\
\And
Rachel Sarig \\
Tel Aviv University, Israel \\
\texttt{sarigrac@tauex.tau.ac.il} \\
\And
Akhilanada Chaurasia \\
King George’s Medical University \\
Lucknow, India \\
\texttt{akhilanandchaurasia@kgmcindia.edu} \\
}

\begin{document}
\maketitle
\begin{abstract}
\textbf{Objectives}: Age and gender estimation is crucial for various applications, including forensic investigations and anthropological studies. This research aims to develop a predictive system for age and gender estimation in living individuals, leveraging dental measurements such as Coronal Height (CH), Coronal Pulp Cavity Height (CPCH), and Tooth Coronal Index (TCI). \\
\textbf{Methods}: Machine learning models were employed in our study, including Cat Boost Classifier (Catboost), Gradient Boosting Machine (GBM), Ada Boost Classifier (AdaBoost), Random Forest (RF), eXtreme Gradient Boosting (XGB), Light Gradient Boosting Machine (LGB), and Extra Trees Classifier (ETC), to analyze dental data from 862 living individuals (459 males and 403 females). Specifically, periapical radiographs from six teeth per individual were utilized, including premolars and molars from both maxillary and mandibular. A novel ensemble learning technique was developed, which uses multiple models each tailored to distinct dental metrics, to estimate age and gender accurately. Furthermore, an explainable AI model has been created utilizing SHAP, enabling dental experts to make judicious decisions based on comprehensible insight.\\
\textbf{Results}: The RF and XGB models were particularly effective, yielding the highest F1 score for age and gender estimation. Notably, the XGB model showed a slightly better performance in age estimation, achieving an F1 score of 73.26\%. A similar trend for the RF model was also observed in gender estimation, achieving a F1 score of 77.53\%. \\
\textbf{Conclusions}: This study marks a significant advancement in dental forensic methods, showcasing the potential of machine learning to automate age and gender estimation processes with improved accuracy. \\
\textbf{Clinical Significance}: 
Accurate age and gender predictions hold importance across diverse domains, including forensic investigations involving both living and deceased individuals. Moreover, beyond its forensic applications, age and gender estimation based on dental measurements holds clinical significance in dental diagnostics and treatment planning.
\end{abstract}


\section{Introduction}
Accurate age and gender estimation using dental features are crucial across various fields, including forensic science and clinical dentistry \cite{avucclu2020determination}. Dental structures, known for their resistance to external factors, are a reliable source for age estimation \cite{de2010age}. Key dental measurements, such as Coronal Height (CH), Coronal Pulp Cavity Height (CPCH), and Tooth Coronal Index (TCI), are closely linked to ageing due to the reduction in dental pulp volume over time \cite{verma2019dental}. Traditional methods of age estimation in forensic sciences using dentition are commonly applied to identify unknown corpses or human remains \cite{willems2002non}. However, these traditional methods sometimes lack precision and reliability. Since much of the research is focused on human remains, these methods typically require the extraction of teeth. Consequently, they are often deemed time-consuming and ethically unsuitable, thereby facing reluctance to adopt. Some recent advancements suggest the use of dental radiographs to measure dental pulp for age and gender determination in the living \cite{machado2023clinical}.\

A widely employed non-destructive approach for age and gender estimation, utilized by both dentists and forensic experts, involves analyzing radiographs of the upper and lower jaw to assess the pulp area \cite{veeraraghavan2010determination}. Although this method does not involve physical alteration of the teeth, it requires the expertise of experienced dental professionals for precise predictions. This study explores a range of machine learning (ML) techniques, which offer a promising approach for age and gender determination in living individuals based on dental pulp measurements due to its ability to handle complex patterns and large datasets. Several ML models, such as  Cat Boost Classifier (Catboost) \cite{najm2023modelling}, Gradient Boosting Machine (GBM) \cite{maalouf2011logistic}, Ada Boost Classifier (AdaBoost) \cite{hastie2009multi}, Random Forest (RF) \cite{biau2012analysis}, eXtreme Gradient Boosting (XGB) \cite{chen2015xgboost}, Light Gradient Boosting Machine (LGB)\cite{fan2019light}, and Extra Trees Classifier (ETC) \cite{sharaff2019extra}, were explored in this study, which
provide diverse capability of capturing nuanced relationships between features and target variables. Additionally, data balancing techniques were adopted to mitigate biases and enhance accuracy. Furthermore, the ML models enable continual refinement through iterative learning, potentially improving accuracy and reliability over traditional methods. This interdisciplinary approach harnesses computational power to advance dental diagnostics, offering a non-invasive and potentially more accurate means of age and gender estimation. \\

Moreover, ensemble learning methods combined with non-trainable combiners were investigated in this study. This approach leverages multiple models' strengths, addressing individual model limitations to enhance overall predictive performance. The primary goals of this study are two-fold: 1) to examine the effectiveness of CH, CPCH, and TCI in estimating age and gender; and 2) to evaluate the performance of ensemble ML models in improving the accuracy and reliability of age and gender estimation. By integrating diverse models, we aim to create a robust and versatile framework for age and gender estimation using dental biometrics. Furthermore, to address the issue of non-interpretable results in ensemble models, eXplainable Artificial Intelligence (XAI) techniques, specifically SHAP (SHapley Additive exPlanations), were adopted in our study, which aims to enhance the interpretability of predictions from the proposed model, empowering dental experts to make more informed decisions by gaining insights into the underlying factors influencing the predictions.\\

The remainder of this paper is organized as follows: Section II presents related work, highlighting existing methods and their limitations. Section III details the proposed methods, including those for data collection, feature extraction, machine learning and ensemble learning for age and gender estimation, and explainable AI (XAI). Section IV provides experimental results and discussions, and finally, Section V concludes the paper, summarizing key findings and outlining avenues for future research.

\section{Related Work}

The estimation of chronological age and gender using dental features has long been a focal point in research, particularly in forensic and anthropological studies. A significant contribution in this area was made by Drusini et al. \cite{drusini1997coronal}, who investigated the correlation between pulp cavity size and individual age. Their method involved measuring the Coronal Height (CH) and Coronal Pulp Cavity Height (CPCH) in molars and premolars of both living individuals and skeletal remains. The method of using the Tooth Coronal Index (TCI) for age estimation, as developed by Drusini et al. \cite{drusini2008coronal}, involves calculating TCI from dental measurements to establish a correlation with the individuals' ages. This innovative approach has proven to be effective for age estimation, demonstrating its applicability to both living individuals and skeletal remains. It underscores the utility of dental measurements, particularly TCI, in forensic and anthropological research. Despite its efficacy, the method proposed by Drusini et al. comes with notable limitations. The accurate calculation of TCI requires specialized expertise, which can limit its accessibility and practicality in diverse settings. Furthermore, the process is not without cost considerations, potentially making it less cost-efficient for widespread use. Additionally, the method was found to lack the significant influence of gender on the TCI, which limits its usefulness in cases where determining gender is also critical.\\

Recognizing these constraints, other researchers have explored automated methods using different dental features. For instance, Farhadian et al. \cite{farhadian2019dental} presented a novel approach using neural networks, focusing specifically on the pulp-tooth ratio in canines for age estimation. This method stands out for its accuracy and non-destructive nature, offering practical benefits in forensic applications. Despite these advancements, the neural network approach also has its challenges. It necessitates a substantial volume of data for effective training and often lacks the level of explainability that is crucial in critical fields such as healthcare and dental forensics. Such limitations highlight the ongoing need for refinement and development of methods in the field of age and gender estimation using dental indicators.\\

Additionally, a contrasting approach to age estimation employs histological analysis of dental pulp, as outlined in a study involving 120 extracted teeth across various age groups \cite{baker2019role}. This method examines changes in odontoblasts, mean vessel area (MVA), mean vessel diameter (MVD), and collagen fiber thickness through histopathological processing, utilizing ANOVA and regression analysis to correlate these parameters with age. Findings indicate a decrease in the number of odontoblasts, MVA, and MVD with age, while collagen fiber thickness increases, suggesting that a combination of these parameters can significantly enhance the accuracy of dental age estimation. Despite its improved accuracy, this technique requires substantial resources, including laboratory equipment and specialized personnel, and incurs higher costs due to the need for reagents for staining and processing. However, its highly invasive nature, necessitating a probe of pulpal tissue, raises ethical concerns, especially for living subjects in forensic contexts. Moreover, the feasibility of preserving pulpal cells after death for analysis remains questionable, further complicating its application in postmortem examinations.\\

A recent investigation introduced a novel Bayesian technique for estimating age, utilizing the volume of the dental pulp chamber determined by three-dimensional cone beam computed tomography (CBCT) images \cite{sironi2018age}. This technique leverages a Bayesian network, a type of probabilistic graphical model, to enhance the precision and reduce bias in age estimation decisions. Demonstrating alignment with forensic standards, this Bayesian approach avoids the common pitfalls of under or over-estimating age and offers more accurate age estimations through the careful consideration of prior probabilities. This approach presents a new direction in dental age estimation, serving as an alternative to the conventional methods employed by regression models and machine learning. However, the application of Bayesian techniques can be demanding computationally, especially with large datasets and intricate models, necessitating substantial processing capabilities and time. Moreover, the effectiveness of this method is closely linked to the availability of CBCT imaging technology, which might not be readily available in all forensic scenarios, potentially limiting its widespread adoption. Furthermore, the ethical considerations surrounding the use of additional X-rays raise questions, especially when alternative methods requiring only periapical X-rays could yield comparable conclusions. This issue prompts a critical evaluation of the justification for employing more sophisticated and potentially harmful X-ray techniques in patient care.\\

In forensic dentistry, the focus has not only been on age estimation but also on gender determination using dental structures. A recent study aimed at evaluating the reliability of gender determination from teeth pulp tissue \cite{veeraraghavan2010determination}. This study involved 60 individuals' teeth (30 male and 30 female) extracted for various reasons. The teeth were divided into three groups, with each group containing an equal number of teeth from males and females. Group 1 consisted of pulp tissue examined immediately after extraction, while Groups 2 and 3 included pulp tissues examined one and five months post-extraction, respectively. The method used involved sectioning the teeth and staining the pulpal cells with quinacrine dihydrochloride, followed by observation under a fluorescent microscope to detect Y chromosome fluorescence in the dental pulp. The results were compelling, with the method achieving high accuracy, specificity, positive predictive value, negative predictive value, and efficiency for freshly extracted teeth and those examined one month post-extraction. This suggests that the fluorescent Y body test is a reliable, simple, and cost-effective technique for gender identification in the immediate post-mortem period of up to one month. However, the method's reliability was confirmed only up to one month after extraction. The efficacy of this method beyond a one-month period remains uncertain. Further, The extraction of teeth for gender determination raises ethical questions, particularly in cases involving living subjects.\\

The research in forensic dentistry, as detailed above, demonstrates a variety of approaches to age and gender estimation using dental structures, each with its unique methods and challenges. Building upon these foundations, our research aims to advance the field by developing an automated method that addresses some of the key limitations observed in previous studies. Our primary goal is to create an automated system that requires fewer expert resources and yields accurate results for both age and gender estimation using dental features. This predictive system is designed to leverage advanced computational techniques, including machine learning methods, to analyze dental characteristics with high precision and minimal human intervention. Furthermore, a significant aspect of our approach is the consideration of ethical implications, particularly regarding the invasiveness of methods used in previous studies. Unlike methods that require tooth extraction, our method is developed with a focus on non-invasive techniques. This decision is driven by the ethical need to preserve the integrity of subjects, especially when dealing with living individuals, and to align with contemporary forensic practices that favour non-destructive analysis. Additionally, our research contributes to the advancement of dental age and gender estimation using ensemble learning techniques, where multiple models are leveraged for distinct dental metrics such as coronal height, pulp cavity measurements, and tooth coronal indices, enhancing accuracy and robustness. Moreover, we integrate explainable AI techniques, specifically SHAP, to provide transparent insights into model predictions, enabling dental experts to make informed decisions while maintaining ethical standards and preserving subject integrity.

\section{Methods}

\begin{figure}[h!]
\centering
\includegraphics[width=0.8\textwidth]{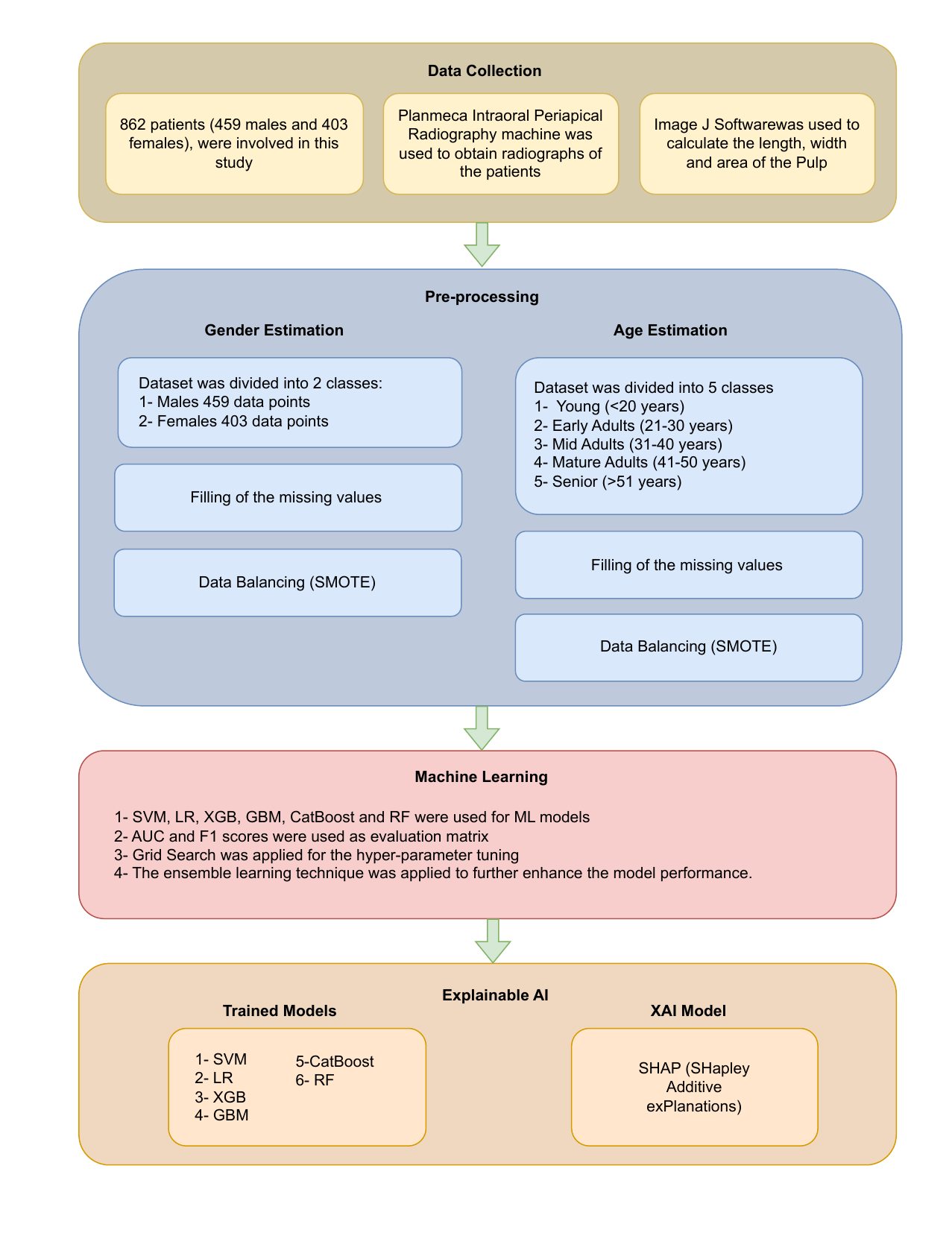}
\caption{Overview of the methodological approach for dental feature-based age and gender estimation, integrating machine learning and explainable AI techniques.}
\label{fig:method}
\end{figure}

An overview of the adopted approach in this study is shown in Figure \ref{fig:method}.

\subsection{Data Collection}

This study adhered to the ethical guidelines of the Helsinki Declaration and received approval from the University of Airlangga, Faculty of Dental Medicine Health Research (Ethical Clearance Certificate Number: 860/HRECC.FODM/X1/2022). Informed consent was obtained from all participants. For this study, we selectively used 862 bitewing radiographs (459 males and 403 females) aged 11 to 70 years (average age 30, median age 27) from the radiology clinic's digital archives. Inclusion criteria were strict: only bitewing radiographs displaying a complete set of maxillary (upper jaw) and mandibular (lower jaw) premolars and molars were selected. Excluded from the study were radiographs showing incomplete teeth sets, distortions, missing teeth, or dental caries. The radiographs were captured using the Planmeca Intraoral Periapical Radiography machine, set at 60Kvp with a 10Amp tube current, and an exposure time of 0.5 seconds. The measurement of pulp length and area was conducted using Image J software \cite{schneider2012nih}, an open-source platform developed at the National Institute of Mental Health in Bethesda, Maryland, USA.

\begin{figure}[h!]
    \centering
    \includegraphics[width=0.8\textwidth]{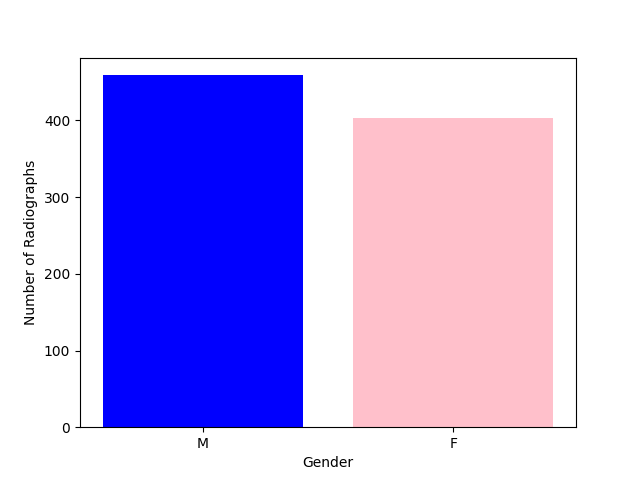}
    \caption{Bar chart showing the distribution of radiographs by gender.}
    \label{fig:gender_radiographs}
\end{figure}

\begin{figure}[h!]
    \centering
    \includegraphics[width=0.8\textwidth]{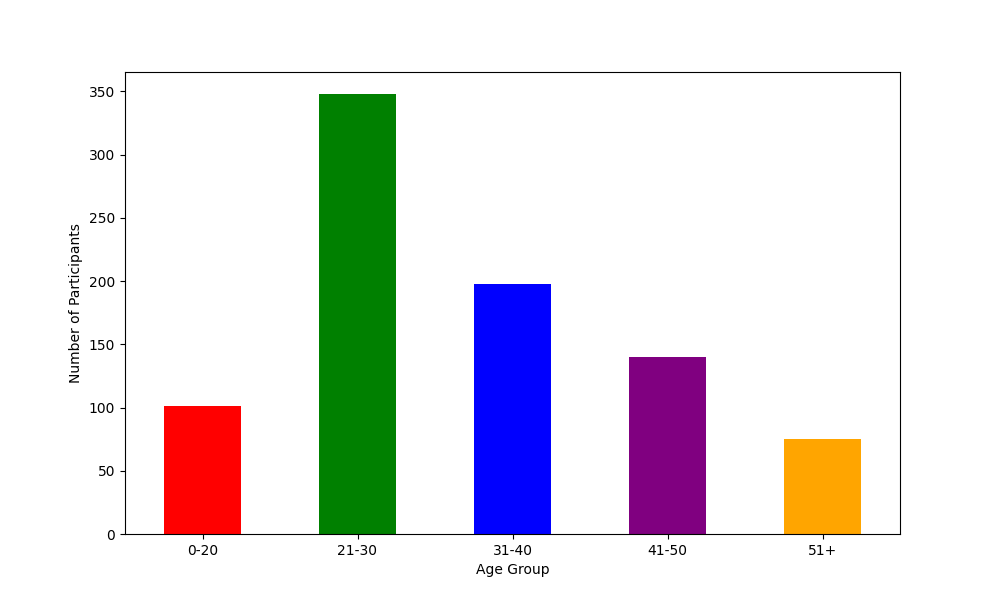}
    \caption{Bar chart showing the distribution of participants across different age groups.}
    \label{fig:age_group_distribution}
\end{figure}

\begin{figure}[h!]
    \centering
    \includegraphics[width=0.8\textwidth]{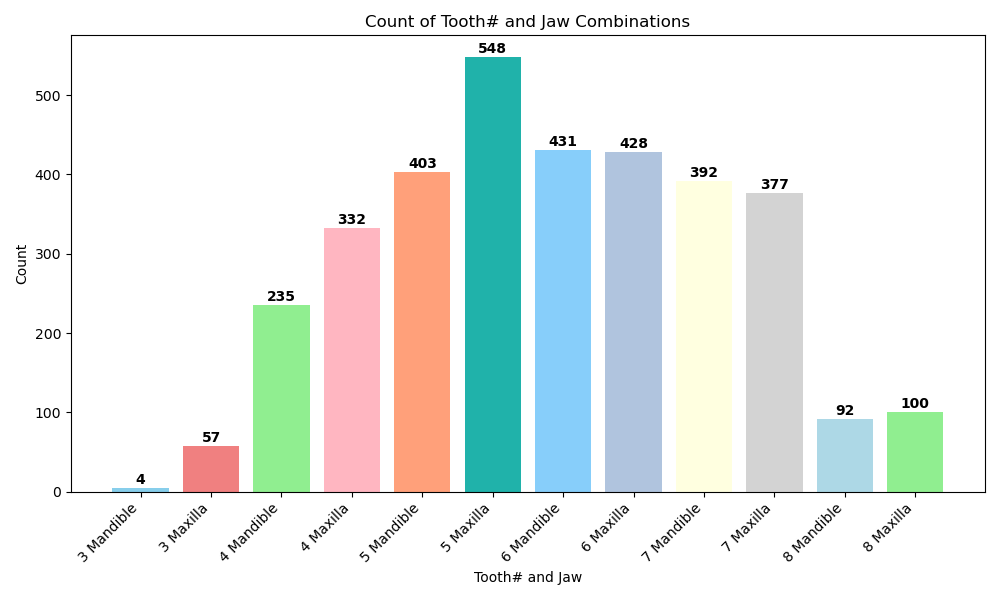}
    \caption{Bar chart showing the frequency of different tooth numbers within the dataset.}
    \label{fig:toothnumber}
\end{figure}

\begin{figure}[h!]
\centering
\includegraphics[width=\linewidth]{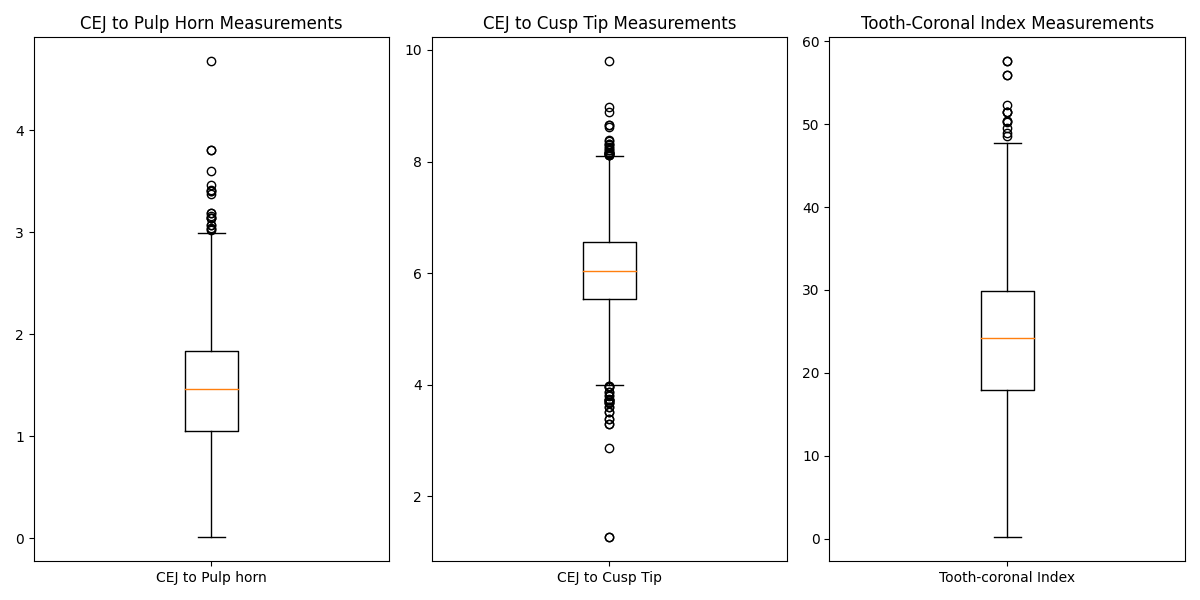}
\caption{Box plot showing the CEJ to Pulp Horn, CEJ to Cusp Tip, and Tooth-coronal Index measurements.}
\label{fig:boxplot}
\end{figure}

For gender estimation, participants were categorized into male and female groups, as shown in Figure \ref{fig:gender_radiographs}. For age estimation, participants were divided into five age groups: Young (0-20 years), Early Adult (21-30 years), Mid Adult (31-40 years), Mature Adult (41-50 years), and Senior (51+ years), as shown in Figure \ref{fig:age_group_distribution}. We allocated 80\% of the dataset for training models, and the remaining 20\% as a test set to evaluate the models' generalization performance for both age and gender estimation.

\subsubsection{Dental Linear Measurements}

Standard adult size 2 (41x31 mm) bitewing radiograph images underwent processing with ImageJ software (version 1.54f; https://imagej.net/)  to convert them into 8-bit grayscale. This conversion reduced the color depth to 256 shades of gray, ranging from 0 (black) to 255 (white) for standardization. Using the film's known dimensions, the images' scale was accurately set within the software. To improve visibility without compromising detail, contrast and brightness adjustments were applied to allow for a slight pixel saturation of 0.35\%. Subsequently, pixels with intensity values below 50 and above 255 were altered to black and white, respectively, which significantly enhanced the visibility of the enamel-dentine junction. Data collection focused on each molar (teeth 6-8), premolar (4-5), and canine (3) visible in the radiographic images from both the upper and lower jaws of each subject. Only teeth free from cavities, excessive wear, diseases, or non-previously treated were included. Measurements of Coronal Height (CH) and Coronal Pulp Chamber Height (CPCH) were obtained using ImageJ by marking four specific points on each tooth: the distal and mesial Cementoenamel Junction (CEJ) as points 1 and 2, the highest point of the pulp horn as point 3, and the tip of the mesio-buccal cusp as point 4 (Figure 6). A line connecting points 1 and 2 was drawn, and the linear distances from this CEJ line to points 3 and 4 were measured to calculate the CH and CPCH (Figure 6). The Tooth Coronal Index (TCI) was then calculated as the ratio of CPCH to CH, multiplied by 100, using the formula:

\begin{equation}
    TCI =  \frac{CPCH\times 100}{CH} 
\end{equation}

\begin{figure}
    \centering
    \includegraphics[width=0.85\textwidth]{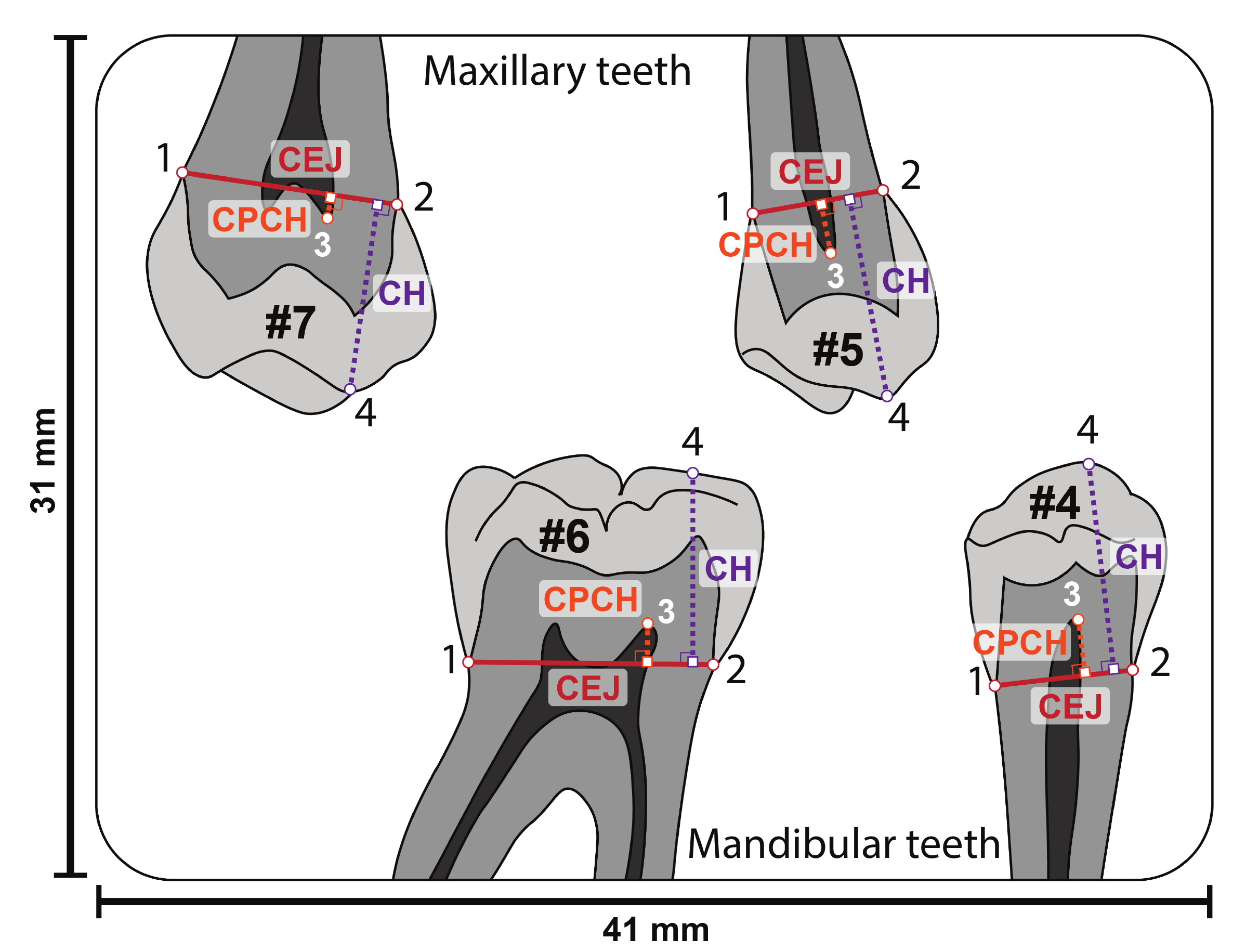}
    \caption{Schematic illustration of teeth measurements carried out on a standard size 2 adult x-ray film. Points 1 and 2 identify the Cementoenamel Junction (CEJ), and the linear distances from the CEJ line to points 3 and 4 are used to determine the Coronal Pulp Chamber Height (CPCH) and Coronal Height (CH), respectively.}
    \label{fig:example}
\end{figure}

\subsection{Data Cleaning and Pre-processing}

In the process of preparing data for analysis, data cleaning and pre-processing are crucial steps to ensure the integrity and quality of the results. Initially, we collected the raw dataset, where each row represents a single tooth from an individual, detailing the tooth's position (maxilla or mandibular), tooth number, and measurements of CH, CPCH and TCI. For this study, we initially considered a range of teeth, including the Canine, First Pre-molar, Second Pre-molar, First Molar, Second Molar, and Third Molar teeth from both mandibular and maxillary sets in the bitewing radiographs provided. However, not all individuals had all these teeth present in the radiographs. Specifically, the Canine, First Pre-molar, and Third Molar teeth were frequently missing, which could introduce bias or inaccuracies in the age and gender estimation models. 

To maintain consistency across the dataset and to ensure the reliability of the predictive analysis, we chose to exclude the Canine, First Pre-molar, and Third Molar teeth from our study. This decision was made to prevent potential data sparsity and to avoid skewing the results with incomplete datasets. Consequently, we focused our analysis on the Second Pre-molar, First Molar, and Second Molar, which were consistently present across the individuals' radiographs. By limiting our study to these teeth, we ensured a more uniform and complete dataset, which is imperative for the accuracy of machine learning models. 

Data pre-processing also involved standardizing the measurements for CPCH and CH across the selected teeth in order to calculate TCI accurately. We reformatted the data so that each row contained comprehensive information for all relevant teeth of a single individual. This reformatting was necessary to align with the input requirements of the machine learning models used in our study, including Catboost, GBM, AdaBoost, RF, XGB, LGB, and ETC. The data after cleaning and pre-processing is described in Table \ref{tab:features_description}

\begin{table}[ht]
\centering
\caption{Features Extracted from Individual Tooth Data in Bitewing Radiographs and Their Descriptions.}
\label{tab:features_description}
\begin{tabular}{|c|l|l|}
\hline
\textbf{No.} & \textbf{Feature} & \textbf{Description} \\
\hline
1 & UniqueID & Identifier for each individual in the dataset \\
\hline
2 & CPCH (mm)\_Second Pre-molar\_Mandible & Distance from CEJ to pulp horn of mandibular Second Pre-molar \\
\hline
3 & CPCH (mm)\_Second Pre-molar\_Maxilla & Distance from CEJ to pulp horn of maxillary Second Pre-molar \\
\hline
4 & CPCH (mm)\_First Molar\_Mandible & Distance from CEJ to pulp horn of mandibular First Molar \\
\hline
5 & CPCH (mm)\_First Molar\_Maxilla & Distance from CEJ to pulp horn of maxillary First Molar \\
\hline
6 & CPCH (mm)\_Second Molar\_Mandible & Distance from CEJ to pulp horn of mandibular Second Molar \\
\hline
7 & CPCH (mm)\_Second Molar\_Maxilla & Distance from CEJ to pulp horn of maxillary Second Molar \\
\hline
8 & CH (mm)\_Second Pre-molar\_Mandible & Distance from CEJ to cusp tip of mandibular Second Pre-molar \\
\hline
9 & CH (mm)\_Second Pre-molar\_Maxilla & Distance from CEJ to cusp tip of maxillary Second Pre-molar \\
\hline
10 & CH (mm)\_First Molar\_Mandible & Distance from CEJ to cusp tip of mandibular First Molar \\
\hline
11 & CH (mm)\_First Molar\_Maxilla & Distance from CEJ to cusp tip of maxillary First Molar \\
\hline
12 & CH(mm)\_Second Molar\_Mandible & Distance from CEJ to cusp tip of mandibular Second Molar \\
\hline
13 & CH (mm)\_Second Molar\_Maxilla & Distance from CEJ to cusp tip of maxillary Second Molar \\
\hline
14 & TCI\_Second Pre-molar\_Mandible & TCI for mandibular Second Pre-molar \\
\hline
15 & TCI\_Second Pre-molar\_Maxilla & TCI for maxillary Second Pre-molar \\
\hline
16 & TCI\_First Molar\_Mandible & TCI for mandibular First Molar \\
\hline
17 & TCI\_First Molar\_Maxilla & TCI for maxillary First Molar \\
\hline
18 & TCI\_Second Molar\_Mandible & TCI for mandibular Second Molar \\
\hline
19 & TCI\_Second Molar\_Maxilla & TCI for maxillary Second Molar \\
\hline
20 & Age & Age of the individual \\
\hline
21 & Gender & Gender of the individual \\
\hline
\end{tabular}
\end{table}

\subsection{Machine Learning Methods for Age and Gender Estimation}

To estimate the age and gender of the individual using radiographic dental features as shown in Table \ref{tab:features_description}, we used several machine learning models including Catboost, GBM, AdaBoost, RF, XGB, LGB, and ETC. These models were selected due to their ability to handle the binary classification for gender estimation and multi-class classification for age group estimation. 

We split the dataset into an 80-20 ratio, with 80\% of data for training the machine learning model and 20\% for validating the model. Moreover, to get generalizable results 5-fold cross-validation was applied. The performance of the machine learning models was not assessed on a single partition of the data but rather on the mean performance across all folds, providing a more reliable estimate of the model's predictive power.  

The primary metric adopted for evaluating the performance of the model is the F1 score, which provides the balance between precision and recall that suit best in this study where data is not balanced. 
In addition to the F1 score, the study also incorporated the Area Under Curve (AUC) performance metric. While AUC is traditionally used in binary classification scenarios, in this study it was adapted for multiclass classification problems. In the multiclass context, AUC is calculated by considering each class against all others, effectively turning a multiclass problem into multiple binary classification problems. This approach, often referred to as the One-vs-Rest (OvR) strategy, involves calculating the AUC for each class separately and then averaging these values to obtain a single AUC score that represents the model's overall ability to distinguish between each class and all other classes. 

Hyper-parameter tuning was done using the grid search, which allows search systematically to select the best combination of the parameters for the enhancement of the model performance.

\subsection{Ensemble Learning}

Ensemble learning \cite{kittler1998combining} is a concept of training multiple models and combining their predictions to increase the performance of the model. In this study, a novel approach was taken to enhance the robustness and performance of the predictive models. Instead of aiming to improve the average results across the folds in K-fold cross-validation, we focused on training distinct models for different teeth, specifically the 5th, 6th, and 7th teeth. Each tooth's unique measurements were utilized to train a separate model, recognizing that different teeth might provide varying levels of significance in age group classification. For each tooth, a dedicated model was trained, harnessing the distinct data distribution pertinent to that particular tooth. This approach acknowledges the individual contribution of each tooth's features to the age group classification task, thereby tailoring the model to leverage the most relevant and informative characteristics. The final stage of the prediction is based on majority voting. In this ensemble method, the predictions from three models (each trained on a different tooth's features) were combined. The final classification for each instance was determined based on the class that received the most votes from these models. By combining the predictions in this manner, the ensemble method leverages the diverse insights provided by each tooth-specific model. This strategy of training separate models for different teeth and then integrating their predictions using majority voting not only enhanced the prediction accuracy but also mitigated the risk of overfitting.

\subsubsection{Non-Trainable Combiners}

Beyond the majority voting that uses the hard output or labels. Non-trainable combiners use the soft output or probabilistic predictions of different architectures. These combiners take advantage of the nuanced probabilities provided by individual classifiers, rather than relying solely on their hard decisions. A Decision Profile Matrix (DPM) is a matrix representation of the probabilistic predictions made by each model in an ensemble for a given test instance. For a classification problem with \( C \) classes and an ensemble of \( N \) models, the DPM for a test instance \( T \) is a \( N \times C \) matrix. Each element \( d_{i,j}(T) \) in the matrix represents the probability assigned by the \( i^{th} \) model to the \( j^{th} \) class for the test instance \( T \). The DPM encapsulates the ensemble's collective knowledge and forms the basis for the operations of non-trainable combiners. In this study, we employed several combiners to determine the best strategy for the given sample data:

\begin{itemize}
    \item \textbf{Mean Combiner}: Calculates the average confidence of the predictions for each class across all models.
    \begin{equation}
        Mean = \frac{1}{N} \sum_{i=1}^{N} p_{i}
    \end{equation}

    \item \textbf{Median Combiner}: Uses the median to determine the central tendency of the predictions, providing robustness against outliers.
    \begin{equation}
        Median = 
        \begin{cases} 
          p_{\left( \frac{N+1}{2} \right)},  \text{if } N \text{ is odd} \\
          \frac{p_{\left( \frac{N}{2} \right)} + p_{\left( \frac{N}{2} + 1 \right)}}{2}, & \text{if } N \text{ is even}
        \end{cases}
    \end{equation}

    \item \textbf{Maximum Combiner}: Selects the maximum confidence prediction for each class from among all models, favouring the most certain predictions.
    \begin{equation}
        Maximum = \max_{i=1}^{N} p_{i}
    \end{equation}

    \item \textbf{Minimum Combiner}: Chooses the minimum confidence prediction for each class, which can be a conservative estimate and beneficial in certain contexts.
    \begin{equation}
        Minimum = \min_{i=1}^{N} p_{i}
    \end{equation}
\end{itemize}

\subsection{Explainable AI (XAI)}

The inherent lack of transparency in complex machine learning models is recognized as a significant limitation, particularly in applications requiring clear understanding and trust in model outputs. To address this challenge, the field of XAI \cite{gunning2019darpa, das2020opportunities, arrieta2020explainable} offers tools and methods to interpret and explain the decision-making processes of these complex ML models. In this section, we employ XAI techniques, specifically focusing on SHAP \cite{NIPS2017_7062}, to analyze the ML models used in this study. As observed in our experiments, RF models demonstrated robust and stable performance in both gender and age estimation tasks. Here, we leverage SHAP to examine the influence of individual dental features on the predictions made by RF models. This exploration through SHAP aims to shed light on how different dental features contribute to the models' decision-making processes in estimating age and gender. Employing SHAP in our analysis not only enhances the interpretability of the ML models but also provides dental experts with deeper insights into the rationale behind each prediction. This added layer of transparency is crucial, particularly in critical fields like healthcare and forensic science, where understanding the 'why' behind model predictions is as important as the accuracy of the predictions themselves. The ability to interpret and validate model decisions is a key step towards wider acceptance and reliance on ML models in these sensitive domains.

\section{Results and Discussion}

\subsection{Performance Evaluation of Machine Learning Models}

This section provides a comprehensive evaluation of the performance of various machine learning models in the context of gender and age estimation. Our approach involved analyzing a range of models, i.e., GBM, Ada Boost, CatBoost, XGB, LGB, RF, and ETC, individually and then collectively through ensemble techniques. The efficacy of these models was measured using two main metrics: F1 score and Area Under the Receiver Operating Characteristic curve (AUC), which offer insights into accuracy and predictive quality.

\begin{table}[h]
    \centering
\caption{Performance Evaluation of ML Models for Gender and Age Classification}
\label{tab:model_performance1}
    \begin{tabular}{cccc}
        \toprule
        \textbf{SNo.} & \textbf{Model} & \textbf{F1} & \textbf{AUC} \\
        \midrule
        \multicolumn{4}{c}{\textbf{Gender Classification}} \\
        \midrule
        1 & Gradient Boosting Classifier & \textbf{63.47} & \textbf{65.74} \\
        2 & Ada Boost Classifier & 62.14 & 65.36 \\
        3 & CatBoost Classifier & 61.05 & 63.67 \\
        4 & Extreme Gradient Boosting & 59.33 & 59.53 \\
        5 & Decision Tree Classifier & 58.83 & 57.19 \\
        \midrule
        \multicolumn{4}{c}{\textbf{Age Classification}} \\
        \midrule
        1 & CatBoost Classifier & \textbf{70.48} & 68.64 \\
        2 & Extra Trees Classifier & 69.41 & \textbf{68.99} \\
        3 & Random Forest Classifier & 68.56 & 68.04 \\
        4 & Light Gradient Boosting Machine & 68.32 & 65.23 \\
        5 & Extreme Gradient Boosting & 66.33 & 64.4 \\
        \bottomrule
    \end{tabular}
\end{table}

Initially to produce the baseline performance described in Table \ref{tab:model_performance1}, we initially trained machine learning models using all available dental features: canine, first pre-molar, second pre-molar, first molar, second molar, and third molar teeth features. Notably, in gender classification, the Gradient Boosting Classifier achieved the highest F1 score of 63.47, suggesting its effectiveness in discerning gender-based features. In contrast, for age classification, the CatBoost Classifier led with an F1 score of 70.48, indicating its robustness in capturing age-related patterns. It's worth noting that as the baseline results were produced without employing ensemble learning techniques, they showcase a decreased performance compared to subsequent analyses incorporating ensemble methods.

We extended our analysis to individual teeth metrics, focusing on teeth numbers second pre-molar, first molar, and second molar, recognizing that certain dental metrics might have varied importance across different age groups. Tables \ref{tab:tooth5_classification}, \ref{tab:tooth6_classification}, and \ref{tab:tooth7_classification} detail the performance of each model for each tooth metric. Across the board, the CatBoost Classifier consistently demonstrated high F1 and AUC scores, emphasizing its adaptability to different dental features for both gender and age classification tasks.

\begin{table}[h]
        \centering
        \caption{Performance Evaluation of ML Models for Gender and Age Classification Using Metrics of Second Pre-molar}
        \begin{tabular}{cccccc}
            \toprule
            \textbf{SNo.} & \textbf{Model} & \textbf{F1 (Second Pre-molar)} & \textbf{AUC (Second Pre-molar)} \\
            \midrule
            \multicolumn{4}{c}{\textbf{Gender Classification}}
             \\
            \midrule
            1 & CatBoost Classifier & 64.88 & 69.61 \\
            2 & Gradient Boosting Classifier & 63.84 & 69.73 \\
            3 & Ada Boost Classifier & 63.7 & 68.62 \\
            4 & Extreme Gradient Boosting & 62.00 & 66.75 \\
            5 & Light Gradient Boosting Machine & 61.74 & 67.12 \\
            \midrule
            \multicolumn{4}{c}{\textbf{Age Classification}} \\
            \midrule
            1 & Extra Trees Classifier & 60.49 & 75.84 \\
        2 & Light Gradient Boosting Machine & 60.16 & 75.83 \\
        3 & Random Forest Classifier & 59.49 & 75.53 \\
        4 & CatBoost Classifier & 59.31 & 75.42 \\
        5 & Extreme Gradient Boosting & 58.61 & 75.1 \\
            \bottomrule
        \end{tabular}
        \label{tab:tooth5_classification}
\end{table}

\begin{table}[h]
        \centering
        \caption{Performance Evaluation of ML Models for Gender and Age Classification Using Metrics of First Molar}
        \begin{tabular}{cccccc}
            \toprule
            \textbf{SNo.} & \textbf{Model} & \textbf{F1 (First Molar)} & \textbf{AUC (First Molar)} \\
            \midrule
            \multicolumn{4}{c}{\textbf{Gender Classification}} \\
            \midrule
            1 & CatBoost Classifier & 66.12 & 70.78 \\
            2 & Gradient Boosting Classifier & 64.91 & 71.15 \\
            3 & Random Forest Classifier & 63.93 & 69.36 \\
            4 & Ada Boost Classifier & 63.7 & 68.74 \\
            5 & Extreme Gradient Boosting & 63.66 & 67.31 \\
            \midrule
            \multicolumn{4}{c}{\textbf{Age Classification}} \\
            \midrule
            1 & CatBoost Classifier & 62.62 & 71.52 \\
            2 & Light Gradient Boosting Machine & 62.03 & 72.18 \\
            3 & Extra Trees Classifier & 61.15 & 71.06 \\
            4 & Extreme Gradient Boosting & 60.78 & 71.89 \\
            5 & Random Forest Classifier & 59.43 & 70.23 \\
            \bottomrule
        \end{tabular}
        \label{tab:tooth6_classification}
\end{table}

\begin{table}[h]
    \centering
        \centering
        \caption{Performance Evaluation of ML Models for Gender and Age Classification Using Metrics of Second Molar}
        \begin{tabular}{cccccc}
            \toprule
            \textbf{SNo.} & \textbf{Model} & \textbf{F1 (Second Molar)} & \textbf{AUC (Second Molar)} \\
            \midrule
            \multicolumn{4}{c}{\textbf{Gender Classification}} \\
            \midrule
            1 & Gradient Boosting Classifier & 64.42 & 67.72 \\
            2 & CatBoost Classifier & 63.7 & 66.11 \\
            3 & Ada Boost Classifier & 62.29 & 65.11 \\
            4 & Decision Tree Classifier & 62.35 & 61.1 \\
            5 & Light Gradient Boosting Machine & 61.49 & 62.24 \\
            \midrule
            \multicolumn{4}{c}{\textbf{Age Classification}} \\
            \midrule
            1 & Extra Trees Classifier & 61.22 & 70.69 \\
            2 & Random Forest Classifier & 61.03 & 70.45 \\
            3 & Light Gradient Boosting Machine & 60.63 & 69.37 \\
            4 & CatBoost Classifier & 59.73 & 69.08 \\
            5 & Extreme Gradient Boosting & 59.43 & 69.16 \\
            \bottomrule
        \end{tabular}
        \label{tab:tooth7_classification}
\end{table}

Furthermore, in extending our research, we adopted an ensemble learning approach to harness the collective strength of multiple models, to improve prediction accuracy and stability. In this methodology, we used different ML models trained on similar features, specifically focusing on teeth numbers 5, 6, and 7. The predictions of these ML models were combined using a majority voting algorithm, where the final output corresponds to the class receiving the highest number of votes from models trained on different teeth. The results, as shown in Table \ref{tab:ensemble_age_classification}, reveal a substantial improvement in both F1 scores and AUC for all models, with Random Forest Classifier and Extreme Gradient Boosting notably achieving the highest scores in gender and age classifications respectively. This enhancement underscores the value of ensemble methods in mitigating individual model biases and variances, particularly in complex classification tasks like age and gender estimation.

Further, we explored the impact of non-trainable combiners, including Mean, Median, Maximum, and Minimum, as detailed in Table \ref{tab:combiners_performance3}. This analysis provided insights into how different combiners affect the performance of various ensemble models. Generally, the Mean Combiner yielded favourable results across the board, suggesting its utility in averaging predictions and mitigating individual model variance. However, certain models benefited more from the Median Combiner, particularly in cases of skewed or unbalanced individual predictions. The Random Forest and Extreme Boosting models consistently showcased robust performance across all combiners, with the Random Forest model excelling in gender classification and Extreme Boosting leading in age classification.

The performance evaluation of various machine learning models and ensemble techniques in this study offers a comprehensive understanding of the potential and challenges in gender and age estimation using dental metrics. The findings underscore the critical role of model selection and the promising potential of ensemble methods in enhancing predictive accuracy. Moreover, the exploration of non-trainable combiners adds a layer of sophistication to understanding ensemble model behaviour, contributing valuable insights for future research and applications in demographic estimation. The robust performance of models like Random Forest and Extreme Boosting, particularly in combination with ensemble techniques, sets a benchmark for future studies aiming to leverage machine learning in the domain of forensic odontology and beyond.

\begin{table}[h]
    \centering
    \caption{Ensemble of Models for Gender and Age Classification}
    \begin{tabular}{cccc}
        \toprule
        \textbf{SNo.} & \textbf{Model} & \textbf{F1} & \textbf{AUC} \\
        \midrule
        \multicolumn{4}{c}{\textbf{Gender Classification}} \\
        \midrule
        1 & CatBoost Classifier & 70.34 & 71.3 \\
        2 & Gradient Boosting Classifier & 69.88 & 69.8 \\
        3 & Ada Boost Classifier & 63.24 & 63.19 \\
        4 & Extreme Gradient Boosting & 76.61 & 76.12 \\
        5 & Random Forest Classifier & 76.54 & 76.1 \\
        6 & Light Gradient Boosting Machine & 75.1 & 75.15 \\
        7 & Ensemble of above models & 77.53 & 77.65 \\
        \midrule
        \multicolumn{4}{c}{\textbf{Age Classification}} \\
        \midrule
        1 & Extra Trees Classifier & 70.61 & 70.55 \\
        2 & Light Gradient Boosting Machine & 72.28 & 71.39 \\
        3 & Random Forest Classifier & 71.23 & 69.25 \\
        4 & CatBoost Classifier & 71.96 & 69.38 \\
        5 & Extreme Gradient Boosting & 73.26 & 71.02 \\
        6 & Ensemble of above models & 73.87 & 71.71 \\
        \bottomrule
    \end{tabular}
    \label{tab:ensemble_age_classification}
\end{table}

\begin{table}[htbp]
\centering
\caption{F1 Score of Using Ensemble Learning with Non-Trainable Combiners for Gender and Age Estimation}
\label{tab:combiners_performance3}
\begin{tabular}{@{}lcccc@{}}
\toprule
Model & Mean Combiner & Median Combiner & Maximum Combiner & Minimum Combiner \\
\midrule
\multicolumn{5}{c}{\textbf{Gender Classification}} \\
\midrule
CatBoost & 70.53 & 69.31 & 69.92 & 67.22 \\
GBM & 70.12 & 69.03 & 68.24 & 67.31 \\
AdaBoost & 63.87 & 61.96 & 61.31 & 60.58 \\
XGB & 75.38 & 74.64 & 73.96 & 71.50 \\
RF & 76.92 & 74.55 & 74.62 & 72.81 \\
LGB & 75.37 & 73.19 & 73.33 & 71.57 \\
\midrule
\multicolumn{5}{c}{\textbf{Age Classification}} \\
\midrule
ETC & 71.13 & 70.82 & 69.92 & 69.22 \\
LGB & 72.10 & 71.03 & 69.24 & 69.31 \\
RF & 71.05 & 68.96 & 69.31 & 67.58 \\
CatBoost & 71.58 & 70.64 & 70.96 & 69.50 \\
XGB & 73.06 & 71.55 & 71.62 & 70.81 \\
\bottomrule
\end{tabular}
\end{table}

\subsection{Explainable AI (XAI)}

\begin{figure}[h!]
\centering
\includegraphics[width=0.65\textwidth]{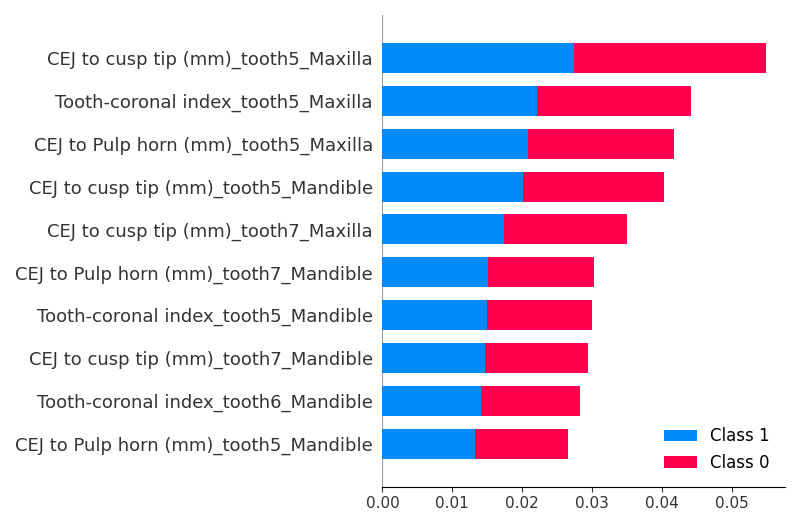}
\caption{SHAP summary plot of top 10 features for gender estimation.}
\label{fig:shapgender}
\end{figure}

Figure \ref{fig:shapgender} shows the top 10 important features for gender classification, indicating that feature "CEJ to cusp tip (mm) Second Premolar Maxilla" significantly influences the model, with "Tooth-coronal index Second Premolar Maxilla" following closely behind. These findings highlight a significant association between these features and the model's ability to estimate gender. The analysis further indicates a dominant role of maxillary dental features over mandibular ones in determining gender. In this plot, the red colour represents the impact of features on female classifications, while blue on male classifications. Observations show that certain features, for example, "CEJ to cusp tip (mm) Second Molar Maxilla", exhibit a noticeable effect on the classification of females, as evidenced by the dominant red shading in its corresponding bar. This dominance suggests that specific dental measurements might be more important for a specific gender. Moreover, analysing the plot it is seen that the Second Premolar Maxilla tooth contains the most important features for the gender classification task.

\begin{figure}[h!]
\centering
\includegraphics[width=0.65\textwidth]
{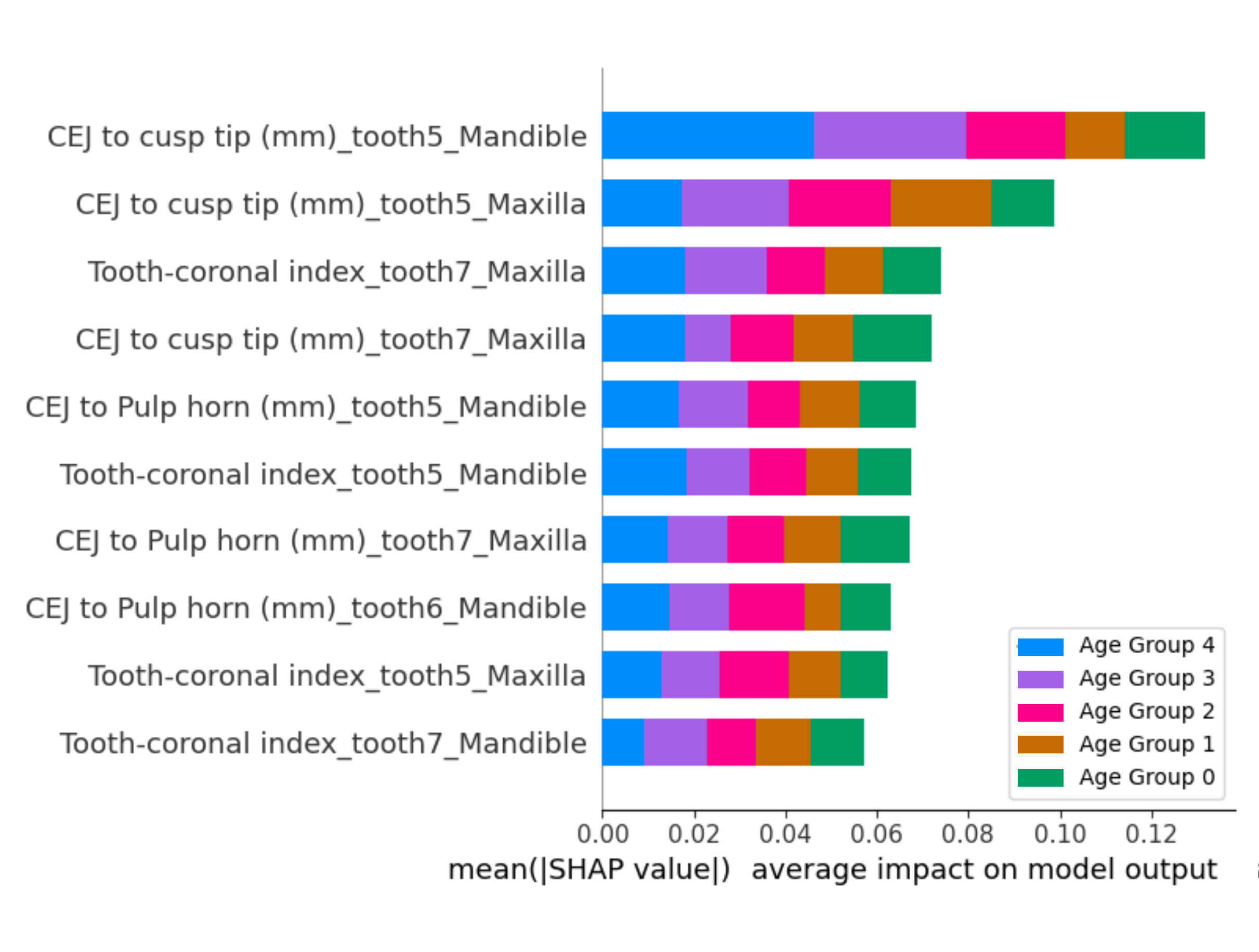}
\caption{SHAP summary plot of top 10 features for age estimation.}
\label{fig:shapage}
\end{figure}

Figure \ref{fig:shapage} shows the influence of the top 10 features on age group classification. Consistent with observations from gender classification, Second Premolar emerges as a significant determinant; however, "CEJ to cusp tip (mm) Second Premolar Mandible" is identified as the most impactful feature for age classification. "Tooth-coronal index" for both Mandible and Maxilla also markedly contributes to the model's predictions, with "Tooth-coronal index Second Molar Maxilla" and "Tooth-coronal index Second Premolar Mandible" being particularly noteworthy. The colour variations within the plot underscore each feature's relevance to certain age brackets, notably Class 0 and Class 4, which represent the youngest and oldest participants, respectively. Unlike the gender classification, the colour distribution across the bars in the age-related plot is not consistent, indicating that various features possess different levels of importance across age groups. This suggests that specific dental measurements may gain or lose predictive power depending on the age class in question, reflecting the progressive changes in dental morphology with ageing.

\subsection{Discussion}

The disparity in model performances between the two tasks, gender and age estimation, underscores the unique challenges inherent in each. Gender estimation, being a binary classification task, often appears less complex, allowing models like Random Forest (RF) and eXtreme Gradient Boosting (XGB) to excel, There appears to be certain features or even dental growth patterns that might align with gender. Other studies simply suggest that sexual dimorphism manifests in different size and shape of the clinical crown \cite{bianchi2023semi} with relatively higher accuracy. Conversely, age estimation, typically treated as a multiclass problem, presents a more intricate challenge, possibly demanding richer feature sets or more advanced modelling techniques, Also as described, one of the most relevant features influencing the performance of the models seems to be the crown height which aligns with clinical observations such as bruxism and overall loss of enamel/dentin structure throughout human lifetime to accurately capture the gradual changes associated with ageing. The consistent improvement in model performance with ensemble methods \cite{pan2019improving} is a key finding of this study, emphasizing the utility of these methods in enhancing predictive accuracy. This improvement is particularly significant in the context of age estimation, where the intricacy of age-related changes requires robust and nuanced model predictions. The use of ensemble methods aligns with the growing trend in machine learning towards leveraging collective model insights. These methods prove to be a powerful tool in complex tasks like age and gender estimation, compensating for the limitations of individual models and providing a more reliable and robust predictive framework. Moreover, the enhanced performance of ensemble models with the introduction of non-trainable combiners, such as Mean, Median, Maximum, and Minimum, indicates their potential to create a harmonious blend of predictions. This approach ensures that the final output is not overly dependent on any single model's biases or weaknesses. Non-trainable combiners, by integrating the strengths of multiple models without additional training, effectively increase the robustness and reliability of the predictions, making them a valuable addition to the ensemble strategies.

\section{Conclusion and Future Work}

In this study, we explored the application of advanced machine learning techniques to the field of forensic science, particularly focusing on age and gender estimation using dental features. Dental characteristics such as CH, CPCH and TCI, which are largely shielded from external environmental effects, were employed as key features for identifying living individuals. This approach marks a significant shift from traditional methods, predominantly used in post-mortem identification, to a more dynamic, living individual-focused analysis which does not require any tooth extraction. This study uses ML models to analyze these dental features representing a substantial advancement, and reducing the reliance on extensive expert analysis and subjective interpretation. Moreover, using dental records which usually include X-rays of some sort, would prove easier and more cost-effective than invasive measures needing laboratory work. Our study effectively utilizes CH, CPCH, and TCI to employ machine learning models in dentistry for estimating an individual's age and gender based on dental biometrics, thereby generating baseline results. This is particularly valuable in forensic contexts where traditional methods may be limited or infeasible due to ethical considerations. A critical element of our methodology was the use of ensemble learning techniques. By integrating the capabilities of various models, ensemble learning ensures that the final predictions are not overly dependent on the quirks of any single model. This approach significantly enhances the reliability and robustness of the predictions. Notably, in the context of age estimation, which presents a more complex multi-class classification problem compared to gender estimation's binary classification, ensemble methods proved particularly effective, improving the accuracy and dependability of the predictions.

While the current study has made significant progress in age and gender estimation using dental features for living individuals, there exists a substantial opportunity to extend this research to the forensic analysis of skeletal remains. This expansion would not only broaden the applicability of the research but also provide invaluable tools in the field of forensic anthropology and archaeology.

\section*{Data Availability}
The dataset and code used in this study are available from the corresponding author upon reasonable request. However, patient consent is required for data disclosure, we may disclose data conditionally through internal discussion and the Institutional Review Board (IRB) of King George's Medical College. 

\section*{Acknowledgment}

The Economic and Social Research Council (ESRC) funded the Business and Local Government Data Research Centre under Grant ES/S007156/1 and supported high-performance computing resources for training the ML models.

\bibliographystyle{plain}
\bibliography{references}

\end{document}